\documentclass[lettersize,journal]{IEEEtran}
\usepackage{amsmath,amsfonts}
\usepackage{algorithmic}
\usepackage{algorithm}
\usepackage{array}
\usepackage{multirow}
\usepackage[caption=false,font=normalsize,labelfont=sf,textfont=sf]{subfig}
\usepackage{textcomp}
\usepackage{stfloats}
\usepackage{url}
\usepackage{verbatim}
\usepackage{graphicx}
\usepackage{cite}
\usepackage{caption}
\usepackage{subcaption}
\usepackage{colortbl}
\usepackage{xcolor}
\usepackage{graphicx}
\usepackage{verbatim}
\usepackage{booktabs}
\begin{document}

\title{Geometry-Aware Cross Modal Alignment for Light Field–LiDAR Semantic Segmentation}

\author{{Jie Luo, Yuxuan Jiang, Xin Jin\textsuperscript{*}~\IEEEmembership{Senior Member,~IEEE}, Mingyu Liu, Yihui Fan}
\thanks{This work is supported by Shenzhen Science and Technology Program under Grant KCXFZ20240903094301003.}
\thanks{Xin Jin is the corresponding author. Jie Luo and Yuxuan Jiang contributed equally to this work.}
\thanks{The authors are with the Shenzhen International Graduate School, Tsinghua University, Shenzhen 518055, China (email:luojie\_tsinghua@163.com; jyx24@mails.tsinghua.edu.cn;jin.xin@sz.tsinghua.edu.cn;
liumingy21@mails.tsinghua.edu.cn; fyh20@mails.tsinghua.edu.cn}}



\maketitle

\begin{abstract}
Semantic segmentation serves as a cornerstone of scene understanding in complex real-world environments. Multimodal sensing, particularly the fusion of light field and LiDAR data, provides complementary appearance and geometric cues, but it remains challenging due to cross-modal discrepancies and limited viewpoint consistency.  To address these challenges, a multimodal semantic segmentation dataset integrating light field data and point cloud data with annotations is presented. 
 Based on this dataset, this paper introduces a multimodal light field–LiDAR fusion network (Mlpfseg) for joint semantic segmentation of images and point clouds. The proposed framework introduces a feature completion module to alleviate the density mismatch between sparse point clouds and dense image representations via differential feature reconstruction, and a depth perception module to enhance occlusion-aware feature learning through attention refinement. Experimental results demonstrate that the proposed method achieves 92.38 mean Intersection over Union (mIoU) on point cloud segmentation and 84.97 mIoU on image segmentation, outperforming existing multimodal methods and improving over the baseline by 2.38 mIoU and 3.75 mIoU, respectively.
\end{abstract}


\section{Introduction}
\IEEEPARstart{A}{s} a fundamental task in computer vision, semantic segmentation plays a key role in applications such as autonomous driving \cite{b1}, road detection \cite{b2}, and medical image analysis \cite{b3}. Existing approaches have evolved from traditional machine learning methods to deep learning-based single-modal and multimodal paradigms.

Early methods relied on handcrafted features and classical machine learning techniques such as clustering and support vector machines. With the success of deep learning, single-modal approaches have become dominant, operating on different sensing modalities, including RGB images \cite{b14}, LiDAR point clouds \cite{Cylinder} and infrared images \cite{infra}. Image-based methods focus on pixel-wise labeling, while LiDAR-based methods predict semantic labels for 3D points.

RGB images provide rich appearance cues but lack 3D geometry and are sensitive to illumination changes, while LiDAR point clouds offer accurate spatial structure but suffer from sparsity and missing texture information. These complementary characteristics have motivated extensive research on multimodal semantic segmentation for improved robustness. However, existing fusion methods such as 2DPASS \cite{b4} and Mseg3D \cite{b5} often rely on asymmetric or weak cross-modal interaction, limiting full exploitation of complementary information. In addition, LiDAR sparsity may even degrade image segmentation performance due to modality imbalance, rather than consistently providing benefits. More importantly, conventional multimodal datasets based on multi-camera systems typically lack sufficient overlapping views, making it difficult to recover occluded regions in complex traffic scenes.

To address this limitation, light field imaging has recently attracted increasing attention. Unlike conventional imaging systems, light field data provides dense multi-view observations with strong viewpoint overlap, enabling more complete perception of occluded objects. UrbanLF \cite{b7} provides rich sub-aperture images but lacks semantic annotations for segmentation. Other light field datasets include annotations but are often constrained by limited baseline and angular resolution, with supervision typically restricted to central views, preventing full exploitation of multi-view information.

To address the issues identified above, we constructed \textit{TrafficScene} \cite{b81}, the first dataset with semantic annotations that includes both light field images and LiDAR point cloud data. Unlike previous datasets, all viewpoints of the light field are annotated, enabling effective information supplementation for occluded and small objects through multi-view consistency.
To effectively integrate light field and point cloud data, we propose a novel fusion-based semantic segmentation framework that jointly performs segmentation on both modalities, thereby better exploiting their complementary characteristics. To alleviate the degradation of light field image segmentation caused by the sparsity discrepancy between point clouds and images, we design a pixel–point feature fusion interpolation module. This module interpolates the features of point clouds projected onto the light field image plane and subsequently fuses them, thus mitigating the negative impact of sparse point clouds on light field image segmentation. To enhance the recognition of occluded objects, we introduce a depth difference perception module, which leverages depth information to perceive occlusions.  

The major contributions are as follows:

1. \textit{TrafficScene},  the first multimodal dataset for semantic segmentation that incorporates light field images and LiDAR point clouds with annotations. Captured using a unique 3×3 camera array with a LiDAR sensor, TrafficScene provides comprehensive semantic annotations across all light field viewpoints, enabling effective multi-view information utilization.

2. We propose a novel light field and LiDAR fusion framework, Multimodal Light field Point Cloud Fusion Segmentation Method (Mlpfseg). It enhances the full integration of point clouds and images and improves the perception of occluded objects through the Point-Pixel Feature Fusion Module and the Depth Difference Perception Module.

3. Extensive experiments demonstrate that Mlpfseg consistently outperforms representative baseline and state-of-the-art methods for single-image, light field image, and multimodal semantic segmentation on the TrafficScene dataset. In particular, it achieves improvements of +3.75 and +2.38 mIoU over the baseline under different settings, while also surpassing previous state-of-the-art methods by up to +1.9 mIoU. Additional results on two benchmark datasets further validate the robustness and generalization capability of the proposed method.

The rest of this paper is organized as follows. Section II summarizes related works on semantic segmentation datasets and semantic segmentation methods. The proposed approach is detailed in Section III. Experiments, including comparisons, ablation studies and visualization are given in Section IV. Finally, we conclude the paper in Section V.

\section{Related Works}
In this section, we introduce existing semantic segmentation datasets, image semantic segmentation methods, light field semantic segmentation methods, point cloud semantic segmentation methods and multimodal fusion semantic segmentation methods.

\subsection{Semantic Segmentation Dataset}
Image-based datasets\cite{b31,r20,r21} typically rely on a single perspective, limiting their ability to capture complete scene information. 
Light field datasets\cite{b7} exploit multi-view information to alleviate occlusion issues; however, their narrow baselines and limited annotation (usually only the central view) restrict their effectiveness. 
Point cloud datasets\cite{r23,r24} provide accurate 3D spatial structure but suffer from sparsity. 
Multimodal datasets\cite{r28,r29,r30,r31,r32} improve performance through cross-modal fusion, yet their reliance on single-view imaging still constrains comprehensive scene understanding.

\subsection{Image Semantic Segmentation Methods}
Early approaches relied on handcrafted features \cite{b13}, while modern methods are dominated by deep learning. 
Representative CNN-based methods, such as FCN \cite{b14}, PSPNet \cite{b15}, and DeepLabV3 \cite{b16}, improve segmentation performance through multi-scale context modeling. 
More recently, transformer-based approaches, including OCRNet \cite{b17}, Mask2Former \cite{b18}, and SegFormer \cite{b19}, further enhance global context modeling and representation capability. 
Despite these advances, single-image methods remain limited in handling occlusions and geometrically ambiguous regions due to the lack of explicit 3D structural information.

\subsection{Light Field Semantic Segmentation Methods}
Light field semantic segmentation leverages both spatial and angular information to improve pixel-level recognition, particularly in occluded or geometrically ambiguous regions. 
Existing methods exploit multi-view consistency and angular cues to enhance feature representation. 
For example, Chen et al.~\cite{b8} employed CNNs with angular modeling and ASPP for multi-scale context extraction, while Sheng et al.~\cite{b7} aggregated sub-aperture images to capture complementary viewpoints. 
Subsequent works incorporated attention mechanisms and depth information \cite{b20}, as well as feature rectification and cross-view fusion \cite{b6}, to improve consistency across views. 
More recently, LF-IENet++~\cite{10440124} introduced an effective feature integration strategy to address multi-baseline disparities. 
However, these methods are fundamentally limited by the narrow baseline of light field cameras, which restricts the available geometric and depth cues.

\subsection{Point Cloud Semantic Segmentation Methods}
Point cloud semantic segmentation aims to assign semantic labels to individual 3D points, leveraging precise geometric structure for scene understanding. 
Existing approaches can be broadly categorized into point-based, projection-based, and voxel-based methods. 
Point-based methods, such as PointNet \cite{b22} and PointNet++ \cite{b23}, directly learn features from raw points using multilayer perceptrons. 
Projection-based methods map point clouds onto 2D representations and apply 2D CNNs \cite{b24,b26}. 
Voxel-based methods discretize space into volumetric grids and employ sparse 3D convolutions, as demonstrated by MinkowskiNet \cite{b27} and SPVCNN \cite{b28}. 
Despite their effectiveness, point cloud methods suffer from inherent sparsity and irregularity, leading to degraded performance in small or occluded object regions.

\subsection{Multimodal Fusion Semantic Segmentation Methods}
Multimodal semantic segmentation integrates complementary information from images and point clouds to improve performance. 
Existing methods can be roughly divided into image-oriented and point cloud-oriented approaches. 
Image-oriented methods, such as CMNeXt \cite{b6}, project point clouds onto the image plane for feature enhancement, but are limited by the sparsity of projected points. 
Point cloud-oriented methods adopt either data-level fusion, e.g., FuseSeg \cite{b29}, or feature-level fusion, such as PMF \cite{b30} and 2DPASS \cite{b4}. 
More advanced approaches, such as MSeg3D \cite{b5}, introduce cross-modal attention to improve fusion effectiveness. 
Nevertheless, most existing methods produce segmentation results for only a single modality, limiting the full exploitation of cross-modal information. 
Moreover, the reliance on single-view images restricts comprehensive scene understanding.

\begin{figure*}[t]
\centering
\captionsetup{labelsep=period}

\subfloat[]{
   \centering
   \includegraphics[width=0.2\linewidth]{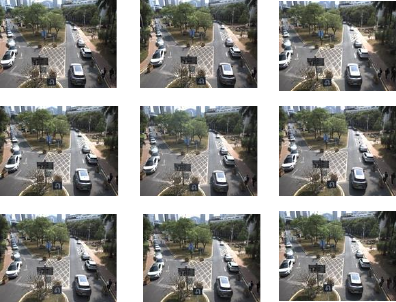}
}
\subfloat[]{
   \centering
   \includegraphics[width=0.185\linewidth]{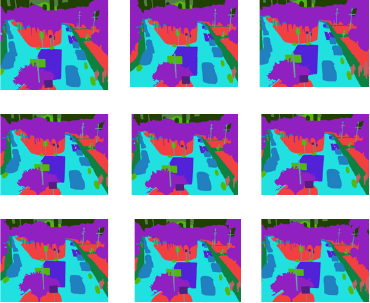}
}
\subfloat[]{
   \centering
   \includegraphics[width=0.183\linewidth]{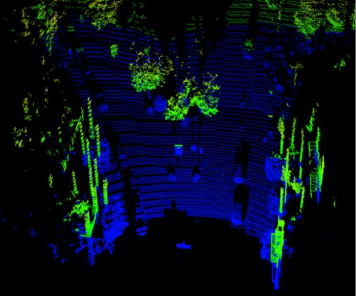}
}
\subfloat[]{
   \centering
   \includegraphics[width=0.181\linewidth]{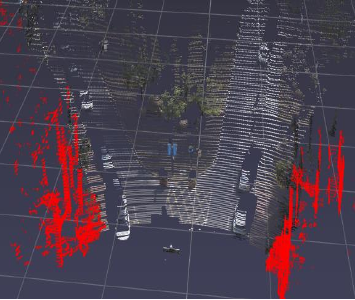}
}
\subfloat[]{
   \centering
   \includegraphics[width=0.182\linewidth]{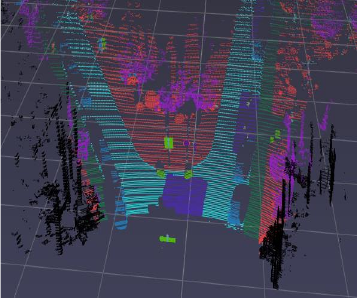}
}

\caption{Multimodal data samples from our dataset. 
(a) Light field images captured by the cameras. 
(b) Pixel-level annotations of the light field images. 
(c) LiDAR point cloud data. 
(d) RGB projection of the point cloud onto the image plane. 
(e) Corresponding annotations of the point cloud data.}
\label{fig:pic1}
\end{figure*}

\begin{figure}[!t] 
    \centering
    \includegraphics[width=0.5\textwidth]{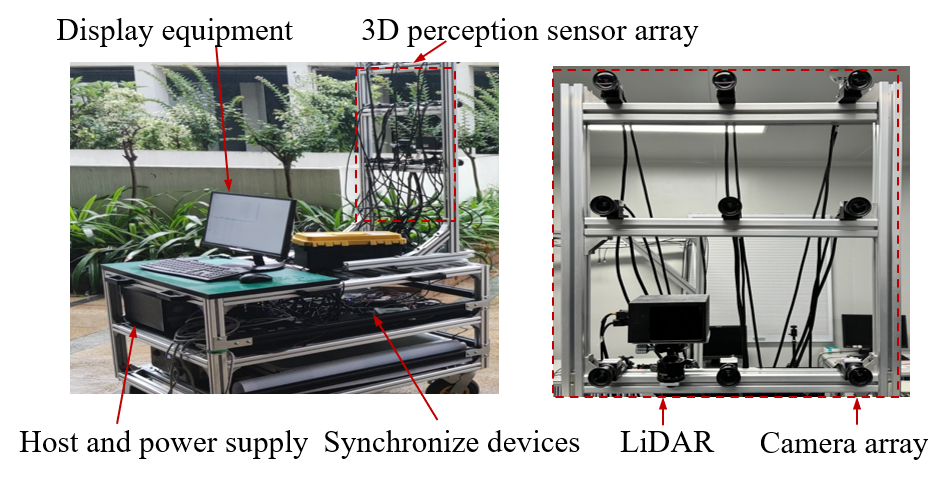} 
    \captionsetup{labelsep=period}
    \caption{Multimodal data acquisition system. }
    \label{fig:equipment}
\end{figure}

\section{Proposed Method}

Building upon our previously established multimodal dataset \textit{TrafficScene}\cite{b81} that integrates light field images and LiDAR point clouds, this paper investigates semantic segmentation in complex outdoor environments. The light field modality provides rich multi-view cues that are effective for handling occlusions, while LiDAR preserves precise 3D geometric structure. Based on this representation, we propose a unified fusion framework, termed Multimodal Light Field Point Cloud Fusion Segmentation (Mlpfseg). The framework consists of two key components: the Point–Pixel Feature Fusion Module (PFFM), which enables fine-grained cross-modal interaction, and the Depth Difference Perception Module (DDPM), which explicitly models cross-modal depth inconsistencies. These designs jointly improve segmentation performance, particularly in occluded regions. The remainder of this section details the proposed framework.

\subsection{Multimodal TrafficScene Dataset}

\begin{figure*}[htbp]  
    \centering
    \includegraphics[width=0.9\textwidth]{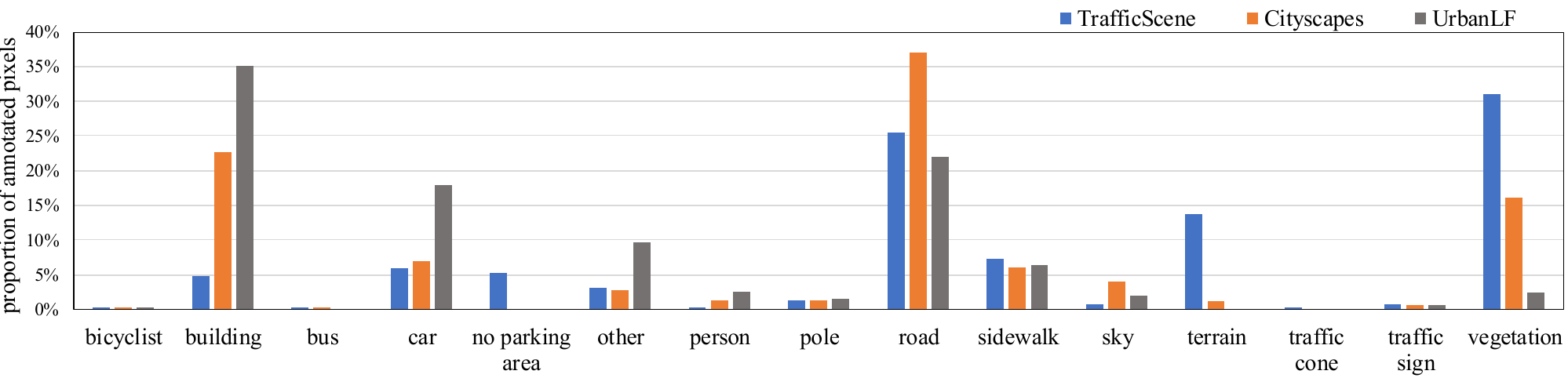}
    \captionsetup{width=\linewidth, justification=centering}
    \captionsetup{labelsep=period}
    \caption{The proportion of annotated pixels (y-axis) per class (x-axis) in TrafficScene, Cityscapes \cite{b31}, UrbanLF \cite{b7}.}
    \label{fig:proportion}
\end{figure*}

TrafficScene is, to the best of our knowledge, the first multimodal dataset that provides jointly annotated LiDAR point clouds and full-view light field images for semantic segmentation. It was originally introduced in our previous work~\cite{b81}, and this paper presents an extended version with improved annotation quality, enhanced cross-modal alignment, and increased diversity of traffic scenarios.

The dataset contains 5607 light field images and 623 LiDAR frames collected using a multimodal acquisition system~\cite{11178181} in real-world traffic environments, as shown in Fig.~\ref{fig:equipment}. The system consists of a 3$\times$3 FLIR BFS-PGE 16S2C camera array~\cite{teledyne2025camera} with a 30 cm baseline and a CH128X1 LiDAR sensor~\cite{leishen2025ch128x1}, enabling accurate synchronization and calibration across modalities. The dataset is available for download at: https://github.com/yxJiang0125/TrafficScene-Dataset

All light field views are distortion-corrected and annotated with 15 semantic categories using CVAT~\cite{cvat2025}, including pedestrians, vehicles, cyclists, traffic signs, roads, sidewalks, vegetation, and other urban elements (Fig.~\ref{fig:proportion}). Cross-view consistency is enforced via label propagation based on feature correspondence and geometric constraints. LiDAR annotations are obtained through calibrated projection and further refined using the Xtreme1 platform~\cite{Xtreme1}. For overlapping projections, semantic labels are merged via majority voting to improve robustness under occlusion.

Overall, the dataset provides synchronized multimodal annotations for both light field images and LiDAR point clouds, enabling comprehensive 2D–3D semantic segmentation. Multi-view light field imagery alleviates LiDAR sparsity, while LiDAR provides complementary geometric structure, forming a strong benchmark for cross-modal scene understanding.
\begin{figure*}[t]
    \centering
    \includegraphics[scale=0.4]{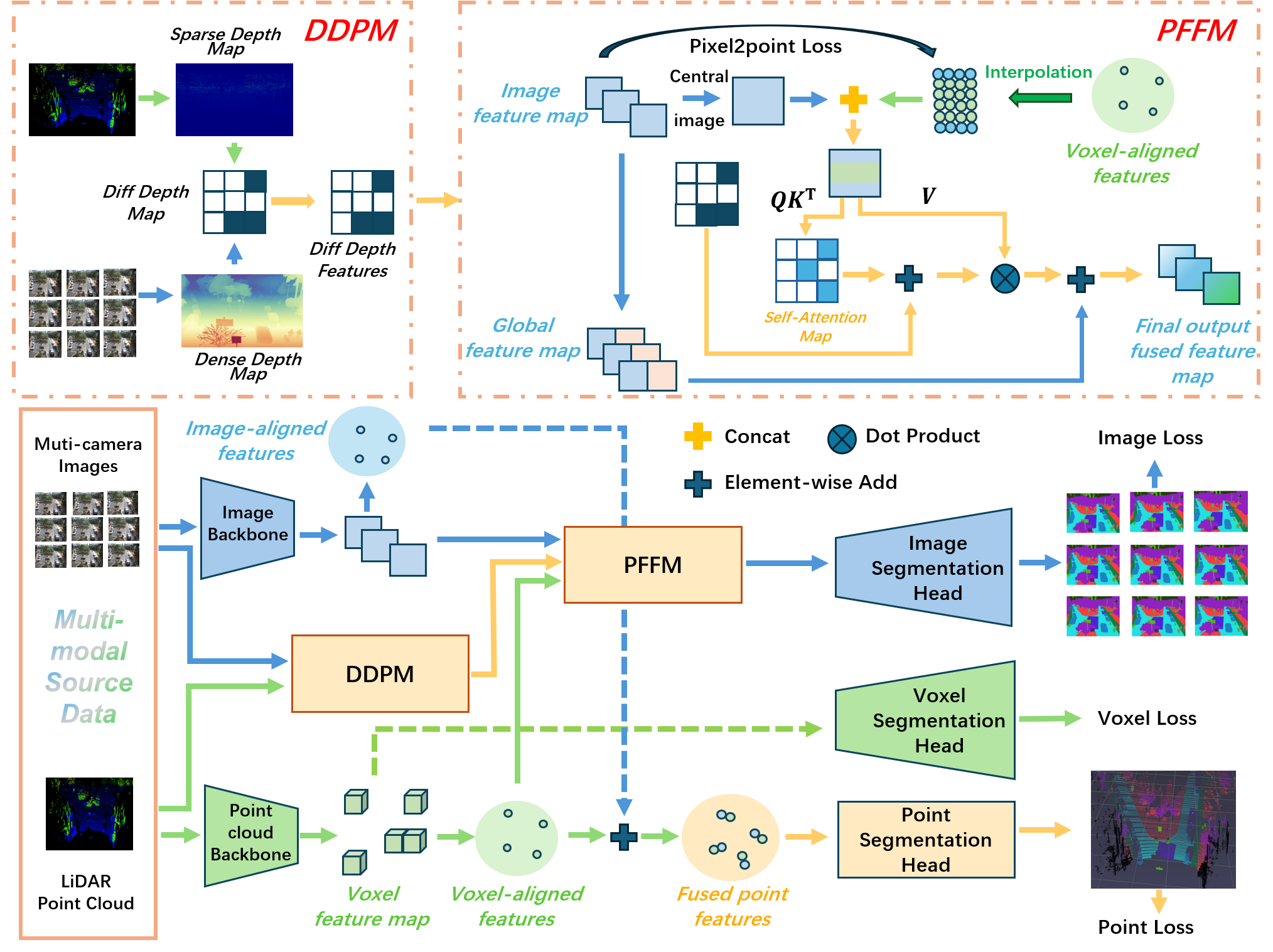} 
    \captionsetup{labelsep=period}
    \caption{Internal structure of multimodal light field point cloud fusion segmentation network. It mainly consists of two parts: point-pixel interpolation fusion module (PFFM) and depth difference perception module (DDPM).}
    \label{fig:network}
\end{figure*}

\subsection{Semantic Segmentation Algorithm Based on Multimodal Data Fusion}

Fig. \ref{fig:network} provides the diagram of the proposed Multimodal Light Field Point Cloud Fusion Segmentation Method (Mlpfseg). Mlpfseg consists of two branches : the light field image branch and the point cloud branch, which are specifically designed for the extraction of image features and point cloud features, respectively. 

For the light field image branch, the input consists of light field images $\{\mathcal{L}_1, \mathcal{L}_2, \cdots, \mathcal{L}_n\}$, where each image has size $\mathbb{R}^{3 \times H \times W}$ and $n$ denotes the number of camera viewpoints. We employ a weight-shared HRNet-48~\cite{sun2019deep} for multi-scale feature extraction, producing viewpoint features $\{F_{\text{img}1}, F_{\text{img}2}, \cdots, F_{\text{img}n}\} \in \mathbb{R}^{c_{\text{img}} \times h \times w}$.

For the point cloud branch, the input point cloud is denoted as $P_{\text{point}} \in \mathbb{R}^{N \times 4}$, where $N$ is the number of points and each point is represented by $\{x_i, y_i, z_i, r_i\}$. The point cloud is voxelized by discretizing coordinates into voxel grids:
\[
Voxel_k = \left\{ \left( \left\lfloor \frac{x_i}{r_l} \right\rfloor, \left\lfloor \frac{y_i}{r_l} \right\rfloor, \left\lfloor \frac{z_i}{r_l} \right\rfloor \right) \right\} \in \mathbb{R}^{N \times 3}.
\]
Voxel features are extracted using SPVCNN~\cite{b28}, yielding $F_{\text{voxel}}^l \in \mathbb{R}^{N_1 \times c_p}$, where $N_1$ denotes the number of non-empty voxels. Point-level features are obtained via interpolation:
\begin{equation}
F_{\text{point}}^l = \sum_{i=1}^{3} \hat{w}_i \cdot F_{\text{voxel}}^l,
\end{equation}
where $\hat{w}_i = \frac{w_i}{\sum_{j=1}^{3} w_j}$ and $w_i = \frac{1}{d(p, v_i) + \epsilon}$. Here $d(p, v_i)$ denotes the distance between point $p$ and its neighboring voxel center $v_i$, and $\epsilon = 10^{-8}$ prevents numerical instability.

After the extraction of the features, the Point-Pixel Feature Fusion Module (PFFM) is proposed to fuse the image features $\{F_{img1}, F_{img2}, \cdots, F_{imgn}\}$ and the features of the voxels $F_{voxel}^i$, which will be discussed in detail in the following subsection \textit{Point-Pixel Feature Fusion Module (PFFM)}. In PFFM, the sparse characteristics of point clouds will be completed, and the fused feature map \(\hat{F}_{fused}\) is obtained.

On this basis, a \textit{Depth Difference Perception Module (DDPM)}  is proposed, with the input predicted depth map for each image $D_{pred}$ and the sparse depth map $D_{sparse}$ presenting the depth values for 3D LiDAR coordinates projected onto the image plane. By utilizing depth difference perception, we obtain the attention score map $\hat{D}_{diff}$ for the occluded objects and send it into the PFFM to optimize the representation in the sparse point cloud module. The detailed description of DDPM will be presented in subsection \textit{Depth Difference Perception  Module (DDPM)}. Ultimately, by inputting the $F_{voxel}^l$ for each layer, Mseg3D \cite{b5}, including multi-scale feature extraction modules and context information fusion modules, and the segmentation head are applied to obtain the fused output $\hat{y}_{img}$ in the image branch and the output $\hat{y}_{point}$ in the point cloud branch.

\subsubsection{Point-Pixel Feature Fusion Module (PFFM)}
After obtaining the point-level features $F_{point}^i$ for the $i$-th point cloud in point cloud branch, we project them onto the image plane. Given the original coordinates $\{x_i, y_i, z_i\}$ of the $i$-th point cloud, the projected coordinates on the image plane are computed as:
\begin{equation}
\begin{bmatrix}
u_i \\
v_i \\
1
\end{bmatrix}^\top = \frac{1}{z_i} \times \mathbf{K} \times \mathbf{T} \times \begin{bmatrix}
x_i \\
y_i \\
z_i \\
1
\end{bmatrix}^\top,
\label{eq:projection} 
\end{equation}
where $\mathbf{K} \in \mathbb{R}^{3 \times 4}$ is the camera intrinsic matrix; $\mathbf{T} \in \mathbb{R}^{4 \times 4}$ is the camera extrinsic matrix. Here, $u_i$ and $v_i$ are the coordinates of the projected point on the image plane obtained through perspective projection. Since the feature map size is smaller than the original image size due to feature extraction by HRNet-48, the corresponding coordinates on the feature map 
are given by \( u_i^\prime = u_i \times \frac{h}{H} \) and \( v_i^\prime = v_i \times \frac{w}{W} \). The projected features on the image plane are denoted as $F_{point}^{img}$.

Considering the projection characteristics of LiDAR point clouds, valid points are mainly concentrated in the central image region, while peripheral areas contain few or no projections. To model this property, we first compute the minimum bounding rectangle $M$ enclosing all projected points on the image plane. The valid projected features within $M$ are denoted as $P_{\text{in}}^{\text{point}} \in \mathbb{R}^{N \times c_p}$, while pixels outside $M$ are initialized with zero.

Specifically, let $(x_{\min}, y_{\min})$ and $(x_{\max}, y_{\max})$ be the minimum and maximum coordinates of the projected points, respectively. The bounding region is defined as $M = [x_{\min}, x_{\max}] \times [y_{\min}, y_{\max}]$.

A grid of coordinates \( (x, y) \) is generated within the rectangle \( M \), where the points with assigned values (i.e., those that have point cloud masks) are labeled as: 
\begin{equation}
\text{mask}(x, y) =
\begin{cases}
1 &  (x, y) \in \{ (x_{indices}, y_{indices}) \} \\
0 & (x, y) \notin \{ (x_{indices}, y_{indices}) \} 
\end{cases}
\label{Eq3}
\end{equation}
The set of assigned points is: \( \{ (x, y) \mid \text{mask}(x, y) = 0 \} \).

For each unassigned point \((x, y)\), we find its three nearest valid points \(\{(x_i, y_i)\}_{i=1}^3\), calculate the interpolation weights, where the weights are inversely proportional to the distance between the unassigned point and 
the valid points \( w_i = \frac{1}{d_i + \epsilon} \), where \( d_i = \sqrt{(x - x_i)^2 + (y - y_i)^2} \). We normalize the weights as \( \hat{w}_i = \frac{w_i}{\sum_{j=1}^{3} w_j} \). The interpolated features are then:
\begin{equation}
\hat{F}_{point}^{img} = \sum_{i=1}^{3} \hat{w}_i \cdot F_{point}^{img}(x_i, y_i) \  
\label{Eq4}
\end{equation}

Thus, the final interpolated feature for \((x, y)\) is:
\begin{equation}
\hat{F}^{img}_{point}(x, y) =
\begin{cases}
\sum_{i=1}^{3} w_i \, F^{img}_{point}(x_i, y_i),
& \text{if } \text{mask}(x, y) = 1 \\

F^{img}_{point}(x, y),
& \text{otherwise}
\end{cases}
\label{Eq5}
\end{equation}

\noindent where 
\begin{equation}
w_i = \frac{d_i + \epsilon}{\sum_{j=1}^{3} d_j + \epsilon}.
\end{equation}

This gives the point cloud projection in the feature map of the image plane within the bounding rectangle \(M\). For regions outside the rectangle, we fill them with the corresponding image feature \(F_{img}\) to obtain the complete point cloud projection feature map filled point \( F_{fill\_point} \) is given by \( F_{img}(x, y) \) if \( (x, y) \notin M \), and \( \hat{F}_{point}^{img}(x, y) \) if \( (x, y) \in M \). 
Since the point cloud and image data features are in different spaces and have different network structures, they are not in the same feature space. To facilitate alignment of the point cloud projection feature map with the image feature space for full fusion, we design an alignment loss function (Pixel2point Loss). This loss function minimizes the difference between the feature spaces of the point cloud and image, enabling better fusion of the two. The alignment loss uses Mean Squared Error (MSE) to measure the difference between the point cloud feature map and the image feature map:
\begin{equation}
\mathcal{L}_{align} = \frac{1}{N} \sum_{(x, y)} \left\| F_{fill\_point}(x, y) - F_{img}(x, y) \right\|_2^2 \  
\end{equation}

We further refine the fused feature map using a self-attention mechanism to capture both intra- and inter-modal dependencies, yielding the final representation $\hat{F}_{\text{fused}}$. Specifically, Query, Key, and Value are projected from $F_{\text{fused}}$ as:
\begin{equation}
Q = F_{\text{fused}} W_Q,\quad
K = F_{\text{fused}} W_K,\quad
V = F_{\text{fused}} W_V,
\end{equation}
where $W_Q \in \mathbb{R}^{C \times C_q}$, $W_K \in \mathbb{R}^{C \times C_k}$, and $W_V \in \mathbb{R}^{C \times C_v}$ are learnable matrices. The attention is computed as:
\begin{equation}
\text{Attention}(Q, K, V) = \text{Softmax}\left(\frac{Q^T K}{\sqrt{C_k}}\right) V^T.
\end{equation}

The attention weights are then applied to the Value to get the final fused feature map \(\hat{F}_{fused}\):
\begin{equation}
\hat{F}_{fused} = \text{LayerNorm}\left(\text{Reshape}\left(\text{Attention}(Q, K, V)\right)\right),
\end{equation}

Finally, the output of the fused feature map \(\hat{y}_{fused}\) is obtained by upsampling \(\hat{F}_{fused}\), and the loss is computed by comparing it with the ground truth:
\begin{equation}
L_{fused\_img} = CE (\hat{y}_{fused}, y_{gt})\ ,
\label{eq:fused_loss} 
\end{equation}

\begin{figure*}[htbp]
    \centering
    \includegraphics[scale=0.5]{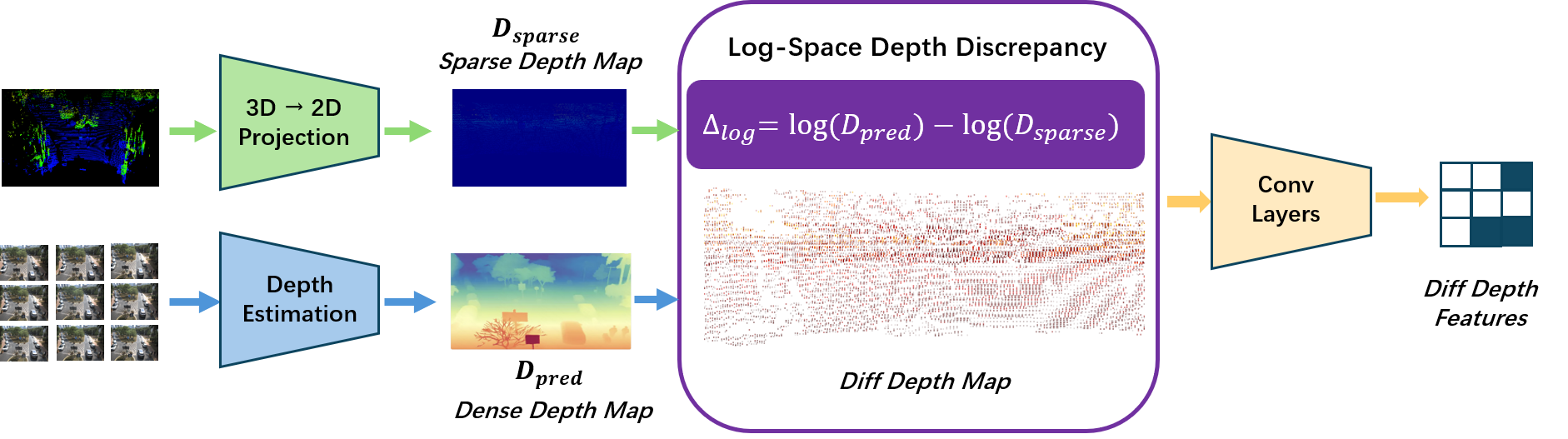} 
    \captionsetup{labelsep=period}
    \caption{Overview of the proposed Depth Difference Perception Module (DDPM). 
The module computes the discrepancy between image-derived depth and point-cloud depth 
to identify occlusion-induced conflicting regions and guides feature fusion accordingly.}
    \label{fig:DDPM}
\end{figure*}

\subsubsection{Depth Difference Perception  Module (DDPM)}
Occlusion introduces inconsistent observations across modalities, where corresponding regions in images and point clouds may contain conflicting geometric cues. A single modality is often insufficient to resolve such ambiguity. To address this issue, we propose a Depth Difference Perception Module (DDPM), as illustrated in Fig.~\ref{fig:DDPM}, which explicitly captures the discrepancy between image-estimated depth and LiDAR depth to identify occluded regions.

\textbf{Depth Discrepancy Modeling.}
Given light field images $\{\mathcal{L}_1, \mathcal{L}_2, \cdots, \mathcal{L}_n\}$, we employ ZoeDepth~\cite{bhat2023zoedepth} to predict dense depth maps $D_{\text{pred}}$. Meanwhile, LiDAR points are projected onto the image plane according to Eq.~(\ref{eq:projection}), yielding sparse depth values $z_i$ at pixel locations $(u_i, v_i)$. This produces a sparse depth map $D_{\text{sparse}}$, where valid entries correspond to projected LiDAR points.

We only consider valid projection locations $\mathcal{V}$. The depth discrepancy is computed as:
\begin{align}
D_{\text{diff}}(i,j) = \log\left(D_{\text{pred}}(i,j) + \epsilon\right) 
- \log\left(D_{\text{sparse}}(i,j) + \epsilon\right),
\end{align}
where $(i,j)\in \mathcal{V}$ and $\epsilon = 10^{-8}$ ensure numerical stability. The logarithmic transformation emphasizes discrepancies at near distances while suppressing noise at far ranges.

\textbf{Feature Enhancement.}
To align the depth discrepancy with the feature space of the network, we apply a lightweight two-layer convolution:
\begin{equation}
\hat{D}_{\text{diff}} = \text{Conv}(\text{Conv}(D_{\text{diff}})).
\end{equation}
The refined discrepancy feature $\hat{D}_{\text{diff}}$ is then integrated into the attention mechanism:
\begin{equation}
\text{Attention}_{\text{final}} = \text{Attention}(Q, K, V) + \hat{D}_{\text{diff}},
\end{equation}
which guides the network to focus on geometrically inconsistent regions caused by occlusion.

\textbf{Loss Functions.}
For the image branch, the segmentation loss is defined as:
\begin{equation}
L_{\text{img}} = CE(\hat{y}_{\text{img}}, y_{\text{gt}}).
\end{equation}

For a single-view input, the total image loss is:
\begin{equation}
L_{\text{img\_total}} =
L_{\text{img}} + L_{\text{img\_lovasz}} + L_{\text{fused\_img}} + \mathcal{L}_{\text{align}}.
\end{equation}

For multi-view inputs, the loss becomes:
\begin{equation}
\begin{aligned}
L_{\text{img\_total}} =
&\; L_{\text{img\_center}} + L_{\text{img\_lvcenter}} + L_{\text{img\_lovasz}} + \mathcal{L}_{\text{align}} \\
&+ L_{\text{fused\_img}}
+ \alpha_1 \sum_{i=1}^{n} L_{\text{img}_i}
+ \alpha_2 \sum_{i=1}^{n} L_{\text{img\_lvi}},
\end{aligned}
\end{equation}
where $L_{\text{img\_lvcenter}}$ denotes the Lovász-Softmax loss applied to the center-view prediction, and $L_{\text{img\_lvi}}$ is its multi-view extension for each sub-view.

For the point cloud branch, the losses are defined as:
\begin{equation}
L_{\text{point}} = CE(\hat{y}_{\text{point}}, y_{\text{point}}) + L_{\text{lovasz}}(\hat{y}_{\text{point}}, y_{\text{point}}),
\end{equation}
\begin{equation}
L_{\text{voxel}} = CE(\hat{y}_{\text{voxel}}, y_{\text{voxel}}) + L_{\text{lovasz}}(\hat{y}_{\text{voxel}}, y_{\text{voxel}}),
\end{equation}
\begin{equation}
L_{\text{point\_total}} = L_{\text{point}} + L_{\text{voxel}}.
\end{equation}

The overall training objective is:
\begin{equation}
L_{\text{total}} = L_{\text{img\_total}} + L_{\text{point\_total}}.
\end{equation}

\section{Experimental Results}
To validate the effectiveness of our proposed the multimodal fusion-based segmentation approach, we conduct  experiments on the \textit{TrafficScene} dataset. All experiments were conducted on a server with an Intel 6330 CPU, 1.0 TB memory, Ubuntu 22.04.5 and CUDA version 12.2. The dataset is split into training, validation, and test sets at a 7:1:2 ratio (3924/594/1116 light field images, 436/66/124 point clouds), using stratified sampling to balance category distributions.

\begin{table*}[!t]
\captionsetup{justification=centering}
\centering
\caption{Quantitative results for image and point cloud semantic segmentation on TrafficScene. Values in parentheses show improvements over previous methods. Red font indicates the state-of-the-art, while blue represents the second-best result.}
\label{tab:final result}
\resizebox{0.9\textwidth}{!}{%
\begin{tabular}{cccc|c|c}
\toprule
Method & Image & Point Cloud & Light Field & Image mIoU & Point Cloud mIoU \\ 
\midrule
FCN \cite{b14}        & $\checkmark$ & $\times$ & $\times$ & 81.23 & -- \\
PSPNet  \cite{b15}    & $\checkmark$ & $\times$ & $\times$ & 81.27 & -- \\
DeepLabV3 \cite{b16} & $\checkmark$ & $\times$ & $\times$ & 80.05 & -- \\ 
OCRNet \cite{b17}  & $\checkmark$ & $\times$ & $\times$ & 82.27 & -- \\
Mask2Former \cite{b18}  & $\checkmark$ & $\times$ & $\times$ & 82.18 & -- \\
SegFormer \cite{b19} & $\checkmark$ & $\times$ & $\times$ & 83.26 & -- \\
PSPNet\_LGA\cite{b81} &$ \times$& $\times$ &  $ \checkmark$ & 81.67 & -- \\
CMNeXt \cite{b6} &$ \times$& $\times$ &  $ \checkmark$ & 83.61 & -- \\
MinkowskiNet \cite{b27}& $\times$ & $\checkmark$ & $\times$ & -- & 84.36 \\
SPVCNN  \cite{b28}    & $\times$ & $\checkmark$ & $\times$ & -- & 85.67 \\
2DPASS  \cite{b4}    & $\checkmark$ & $\checkmark$ & $\times$ & -- & 70.89 \\
PMF     \cite{b30}    & $\checkmark$ & $\checkmark$ & $\times$ & -- & 74.96 \\
Mseg3D   \cite{b5}   & $\checkmark$ & $\checkmark$ & $\times$ & -- & 90.00 \\
DGFusion \cite{DGfusion}    & $\checkmark$ & $\checkmark$ & $\times$ &  -- & 90.45 \\
\midrule 
Baseline    & $\checkmark$ & $\checkmark$ & $\times$ & 81.32 & 90.00 \\
Mlpfseg (one view) & $\checkmark$ & $\checkmark$ & $\times$ & \textbf{\textcolor{red}{85.23}}(\textbf{\textcolor{red}{+3.91}})  & \textbf{\textcolor{blue}{91.50}}(\textbf{\textcolor{blue}{+1.50}}) \\
Mlpfseg (light field images) & $\times$ & $\checkmark$ & $\checkmark$ &  \textbf{\textcolor{blue}{84.97}}(\textbf{\textcolor{blue}{+3.75}}) & \textbf{\textcolor{red}{92.38}}(\textbf{\textcolor{red}{+2.38}}) \\
\bottomrule
\end{tabular}%
}
\end{table*}

\begin{figure}[!t] 
    \centering
    \includegraphics[width=0.45\textwidth]{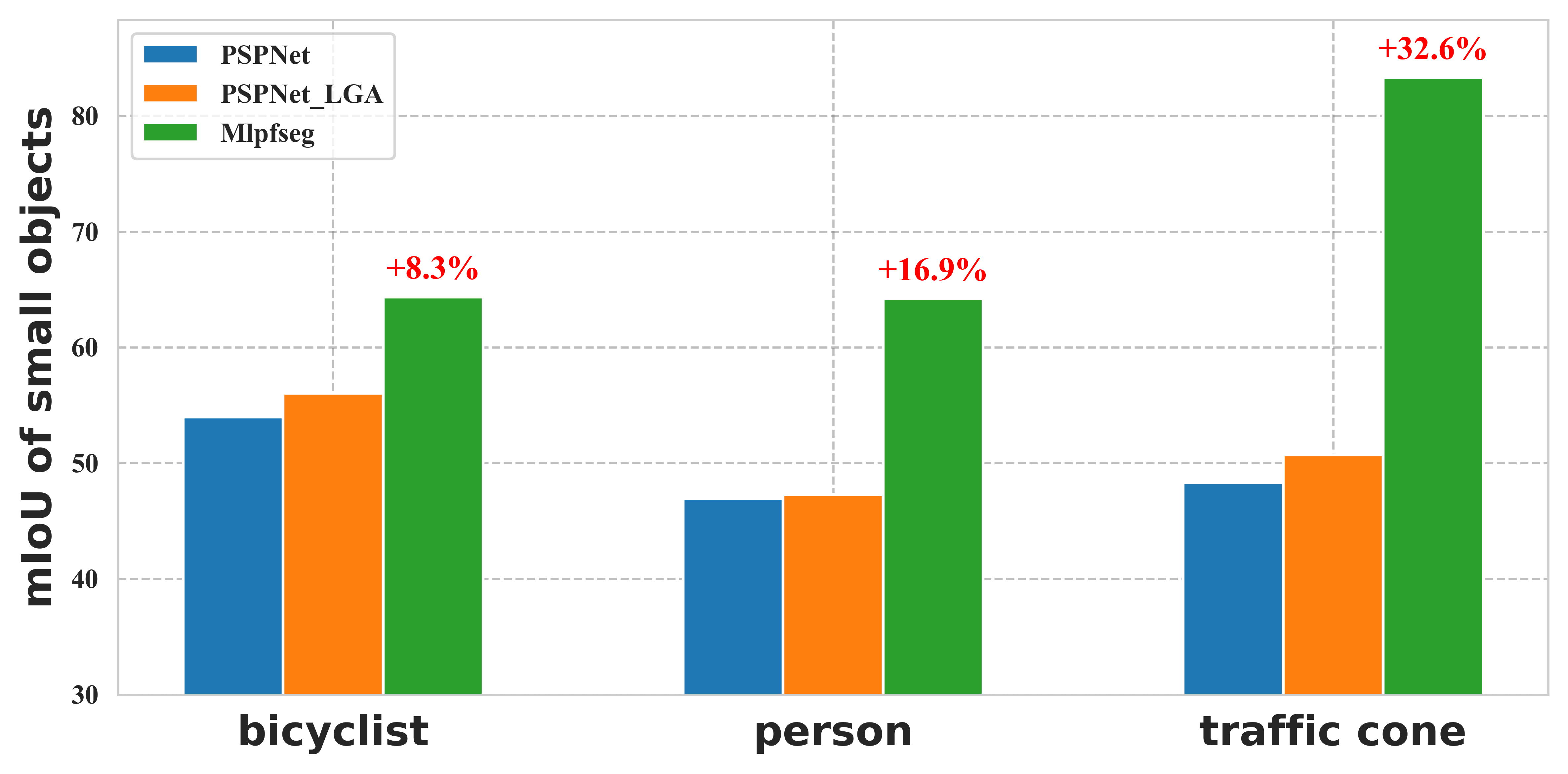} 
    \captionsetup{labelsep=period}
    \caption{mIoU for PSPNet, PSPNet\_LGA and Mlpfseg on small objects across all viewing angle}
    \label{fig:picvis}
\end{figure}

\begin{figure*}[t]
   \centering
   \captionsetup{labelsep=period}

   \begin{minipage}{\linewidth}
      \centering
      \subfloat{
         \includegraphics[width=0.9\linewidth]{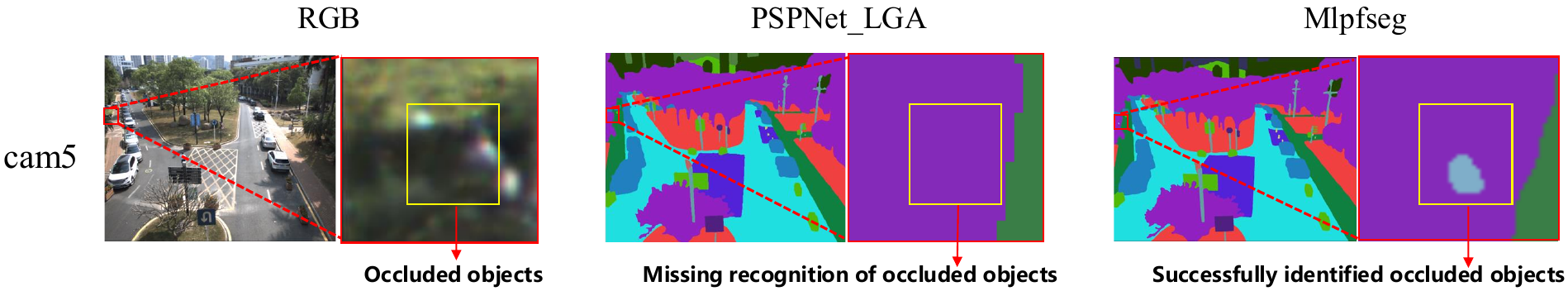}
      }
   \end{minipage}

   \vspace{2mm}

   \begin{minipage}{\linewidth}
      \centering
      \subfloat{
         \includegraphics[width=0.9\linewidth]{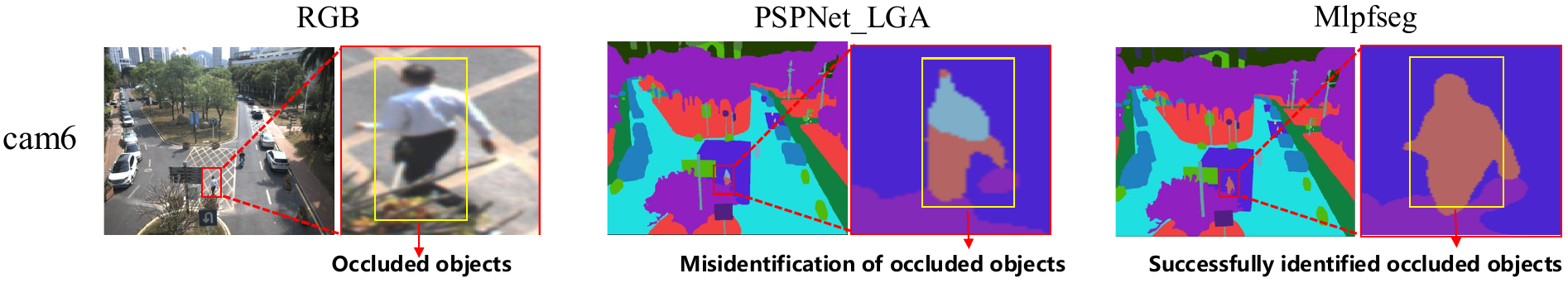}
      }
   \end{minipage}

   \caption {comparison of occluded object segmentation in challenging traffic scenes. The top example shows an occluded bicycle, where PSPNet\_LGA~\cite{b81} fails to recognize the target object under heavy occlusion. The bottom example shows an occluded pedestrian, where PSPNet\_LGA suffers from incorrect identification, while the proposed Mlpfseg produces robust and accurate segmentation results.}
   \label{fig:picnn}
\end{figure*}

\begin{figure*}[!t] 
    \centering
    \includegraphics[width=0.9\textwidth]{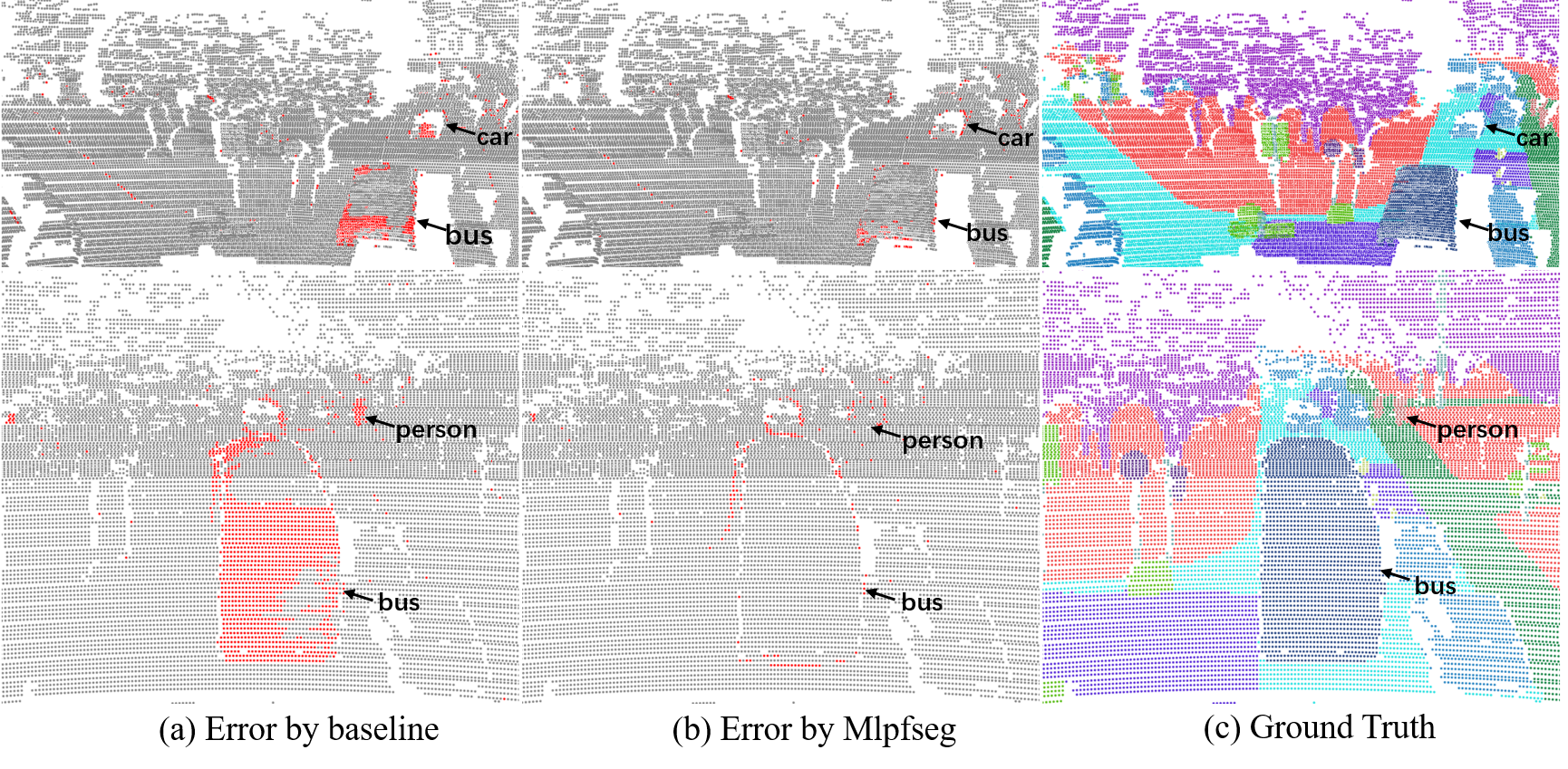} 
    \captionsetup{labelsep=period}
    \caption{Qualitative results of Mlpfseg on the test set of TrafficScene. Our baseline has a higher error recognizing small objects and partially occluded objects.}
    \label{fig:pointcloud}
\end{figure*}

\subsection{Semantic segmentation algorithm based on multimodal data fusion}

\subsubsection{Implementation Details}

\textbf{Image Branch:} Input images of size $1080\times1440$ are augmented with random horizontal flipping (0.5), color jittering, JPEG compression noise (quality [30, 70]), and random cropping (60\%–75\%). We adopt HRNet-W48\cite{sun2019deep} as the backbone, initialized with ImageNet pre-trained weights, and freeze the first three stages during training. Multi-scale features from four stages are fused to obtain $F_{\text{img}} \in \mathbb{R}^{c_{\text{img}} \times h \times w}$ with $c_{\text{img}}=48$, followed by an FCN-based segmentation head. Three camera views are used to provide complementary spatial information.
\textbf{Point Cloud Branch:} We use a modified UNet3D with voxel-based encoding. Point clouds are voxelized within $[x,y,z]\in[-50,6,-7]$ to $[50,106,11]$ with voxel size 0.1\,m and a maximum of 5 points per voxel. The network adopts an encoder–decoder structure with 8$\times$ down-/up-sampling and channel scaling factor 2, producing voxel features $F_{\text{voxel}}^l \in \mathbb{R}^{N_1 \times c_p}$ where $c_p=48$. We follow MSeg3D~\cite{b5} for remaining settings.
\textbf{Training Details:} The model is trained end-to-end using Adam optimizer with an initial learning rate of $2\times10^{-4}$ and weight decay of 0.01. A one-cycle learning rate schedule is adopted with momentum ranging from 0.95 to 0.85. Training is performed on a single NVIDIA A40 GPU for 24 epochs with batch size 1, which is also used during inference.
\subsubsection{Experimental Results}
To validate the effectiveness of our dataset and multimodal fusion method, we conducted extensive experiments. First, we assessed the dataset’s validity by applying established image and light-field semantic segmentation methods, evaluating performance using mean intersection over union (mIoU). We conduct extensive evaluations against representative methods across different modalities, including image-based models (e.g., PSPNet~\cite{b15}, DeepLabV3~\cite{b16}, SegFormer~\cite{b19}), light-field-based methods (PSPNet\_LGA~\cite{b81}, CMNeXt~\cite{b6}), LiDAR-only methods (SPVCNN~\cite{b28}, MinkowskiNet~\cite{b27}), and multimodal fusion approaches (PMF~\cite{b30}, 2DPASS~\cite{b4}, MSeg3D~\cite{b5}, DGFusion~\cite{DGfusion}). All methods are evaluated under identical settings for a fair comparison, and results are reported in Table~\ref{tab:final result}.

The results show that attention-based models consistently outperform convolution-based ones in single-image segmentation. Among them, SegFormer~\cite{b19} achieves the best performance with 83.15 mIoU. For light-field segmentation, CMNeXt~\cite{b6} further improves performance by leveraging multi-view sub-aperture information, reaching 83.61 mIoU. For LiDAR-based segmentation, DGFusion~\cite{DGfusion} achieves the best performance among all compared methods, demonstrating the effectiveness of image-guided fusion. In contrast, projection-based methods such as PMF~\cite{b30} suffer from 3D structural information loss, resulting in inferior performance compared to point-cloud-only methods such as SPVCNN~\cite{b28} and MinkowskiNet~\cite{b27}.

Our proposed Mlpfseg consistently outperforms existing state-of-the-art methods in both image and point cloud segmentation, benefiting from image feature interpolation, an attention-based image encoder, and the proposed Depth Difference Perception Module (DDPM) for occlusion modeling. The performance gains can be attributed to three key factors: image feature interpolation alleviates the sparsity caused by point cloud projection, the attention-based image encoder enhances discriminative feature extraction, and the DDPM improves occluded object modeling by explicitly capturing cross-modal depth discrepancies.

As shown in Fig.~\ref{fig:picvis}, Mlpfseg achieves significant improvements on small and thin objects such as bicyclists, pedestrians, and traffic cones. The integration of multi-view and multimodal information provides more complete contextual and geometric cues, leading to more accurate segmentation of small objects. Fig.~\ref{fig:picnn} demonstrates that our method effectively handles occluded objects, significantly reducing both missed detections and misclassifications compared with light-field-based methods. This improvement is largely attributed to the proposed DDPM, which enhances the network's ability to perceive occlusion relationships across modalities. The point cloud results in Fig.~\ref{fig:pointcloud} show that Mlpfseg produces more accurate predictions under challenging scenarios such as partial vehicle occlusion and small object segmentation. The reduction of mispredicted regions highlighted in red further confirms the effectiveness of the proposed multimodal fusion strategy.

\subsubsection{Generalization on Public Datasets}
To evaluate generalization, we conduct experiments on two standard benchmarks, SemanticKITTI~\cite{semanticKitti} and nuScenes~\cite{nuscenes}. SemanticKITTI provides 22 sequences with 19 semantic classes, while nuScenes contains 1000 driving scenes with RGB–LiDAR data and 17 categories.

Following standard protocols, we report point cloud mIoU as the evaluation metric. Since these datasets do not provide light-field inputs, Mlpfseg is evaluated using a single RGB view together with LiDAR for a fair comparison with existing RGB–LiDAR methods. The results are summarized in Table~\ref{tab:sota_lidar}. 

\begin{table}[!t]
\captionsetup{justification=centering}
\centering
\caption{Comparison with state-of-the-art methods on the SemanticKITTI and nuScenes datasets.}
\label{tab:sota_lidar}
\begin{tabular}{ccc}
\toprule
\textbf{Method} & \textbf{SemanticKITTI\cite{semanticKitti}} & \textbf{nuScenes}\cite{nuscenes} \\
\midrule
SPVCNN \cite{b28} & 66.4 & 77.3 \\
Cylinder3D \cite{Cylinder}  & 64.9 & 76.1 \\
2DPASS \cite{b4} & 70.1 & 80.5 \\
PMF \cite{b30} & 63.9 & 76.0 \\
Mseg3D \cite{b5} & 70.4 & 81.1 \\
PASeg \cite{PASeg} & 70.5 & 80.5 \\
\midrule
Baseline & 67.9  &  80.5 \\
Mlpfseg (one view) & 71.8 & {81.6} \\
\bottomrule
\end{tabular}
\end{table}

\begin{table}[h]
\captionsetup{justification=centering}
\centering
\caption{Ablation experiment}
\label{tab:ablation}
\begin{tabular}{cccc}
\toprule
\multirow{2}{*}{Method} & \multicolumn{3}{c}{mIoU Results} \\
\cmidrule(lr){2-4} 
 & Image & Point Cloud & Average \\
\midrule
Baseline & 81.32 & 90.00 & 85.66 \\
+Aligned Loss & 82.90 & 89.84 & 86.37 (+0.71) \\
+Interpolation Attention Feature & 84.75 & 90.48 & 87.62 (+1.96) \\
+Depth Map & 85.23 & 91.50 & 88.37 (+2.71) \\
+Light Field Image & 84.97 & 92.38 & 88.68 (+3.02) \\
\bottomrule
\end{tabular}
\end{table}

\subsection{Ablation}
Table~\ref{tab:ablation} presents the ablation study results. The baseline extends MSeg3D by incorporating the image branch. Introducing the alignment loss improves image segmentation but slightly affects point cloud performance due to partial cross-modal misalignment, yielding an overall gain of 0.71 mIoU.

The proposed interpolation-based feature interaction significantly improves image segmentation and also benefits point cloud performance, leading to a 1.96 mIoU increase. Incorporating the depth difference perception module further enhances occluded object modeling, resulting in a 2.71 mIoU gain. Finally, integrating multi-view light-field images achieves the best performance with an additional improvement of 3.02 mIoU.

Overall, the results validate the effectiveness of the proposed interpolation, depth perception, and multi-view fusion components.

\section{CONCLUSION}
In this work, we introduce, to the best of our knowledge, the first multimodal dataset for real-world traffic scenes that jointly integrates light field images and LiDAR point clouds. The dataset comprises 623$\times$9 synchronized frames captured by a 3×3 camera array, providing dense multi-view light field observations together with corresponding LiDAR point clouds. A key characteristic of TrafficScene is its comprehensive 2D–3D joint supervision, achieved through dense per-view light field annotations and geometrically consistent LiDAR point-level labels.
Based on this dataset, we benchmark representative single-image, light-field, and point-cloud segmentation methods, demonstrating the effectiveness of multimodal learning for complex traffic scene understanding.
We further propose Mlpfseg, a multimodal fusion framework that jointly processes light field images and LiDAR point clouds. By exploiting cross-modal feature interactions and depth-aware representation learning, the proposed method enhances feature discriminability and improves segmentation performance, particularly for small and occluded objects, achieving consistent gains across both modalities.
In future work, we aim to incorporate light field depth estimation for joint optimization, further improving geometric consistency and enabling a more unified multimodal learning framework.


\vspace{-38pt}
\begin{IEEEbiography}[{\includegraphics[width=1in,height=1.25in,clip,keepaspectratio]{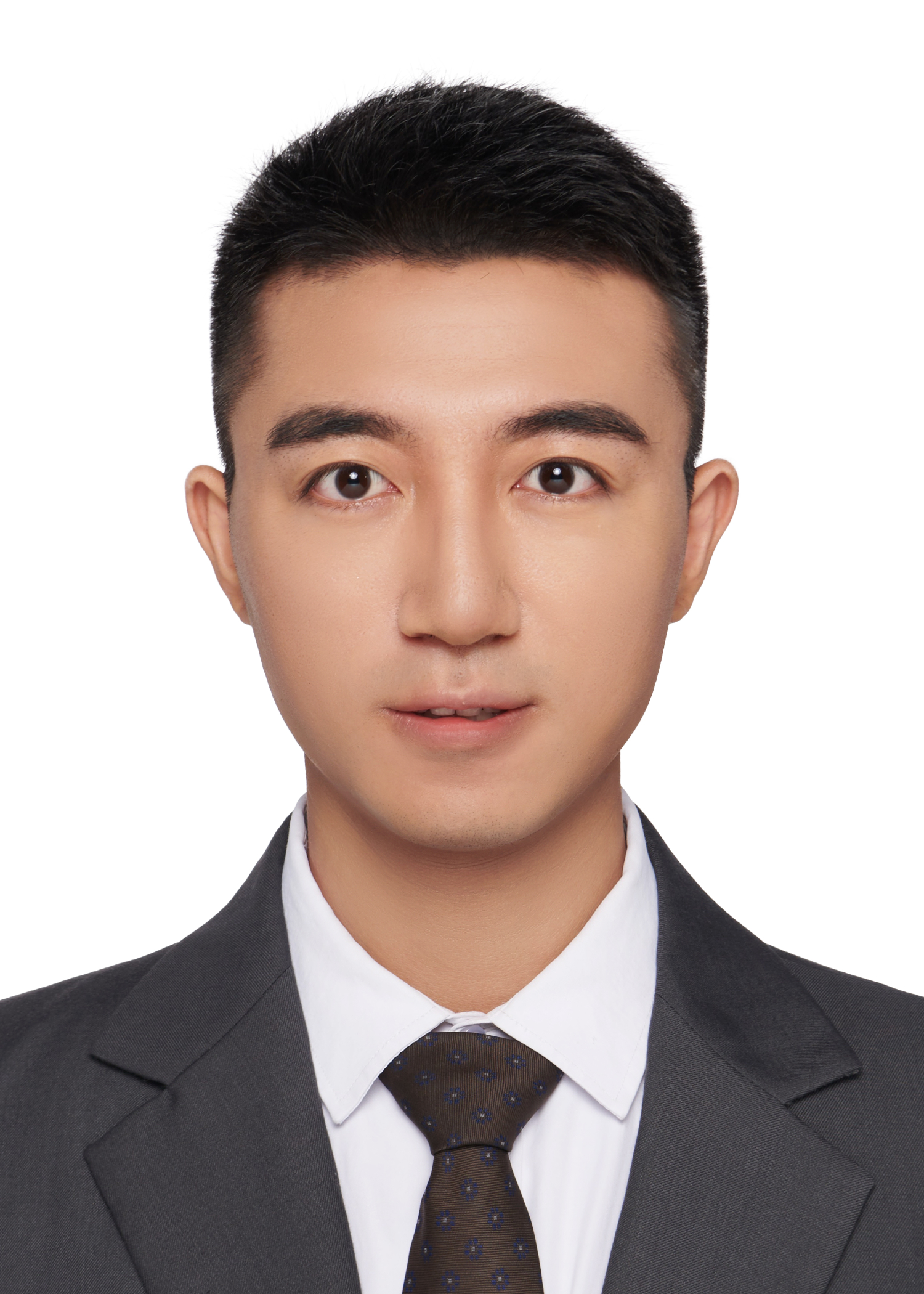}}]{Jie Luo} is currently working toward the Master degree in the Big Data Technology and Engineering with Shenzhen International Graduate School, Tsinghua University, China. His research interests include the autonomous driving, semantic segmentation, and multimodal data fusion. He has published paper in ICME.
\end{IEEEbiography}
\vspace{-38pt}
\begin{IEEEbiography}
[{\includegraphics[width=1in,height=1.25in,clip,keepaspectratio]{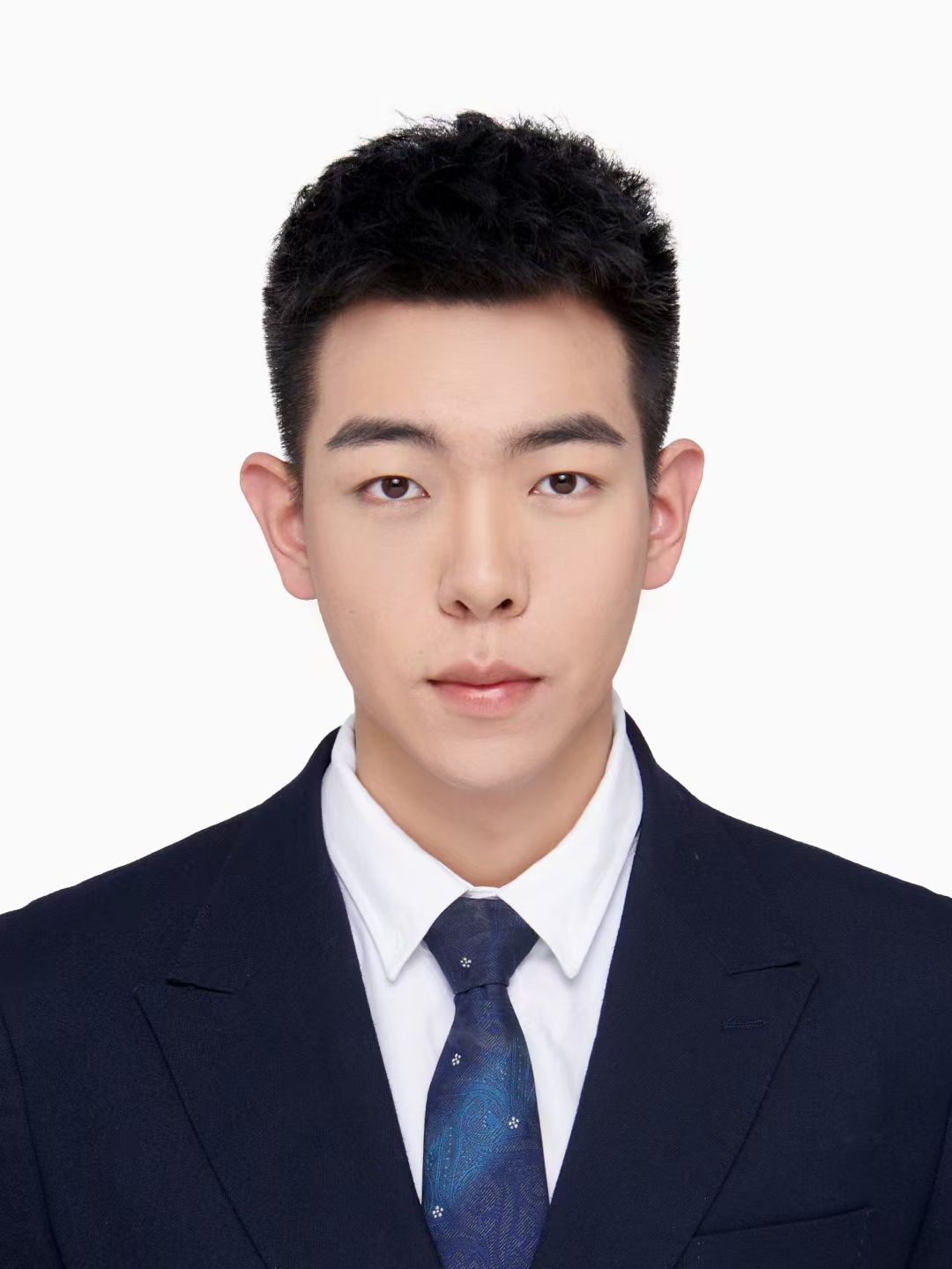}}]{Yuxuan Jiang} is currently working toward the Master degree in the Artificial Intelligence with Shenzhen International Graduate School, Tsinghua University, China. His research interests include the video anomaly detection and multimodal data fusion.
\end{IEEEbiography}
\vspace{-38pt}
\begin{IEEEbiography}[{\includegraphics[width=1in,height=1.25in,clip,keepaspectratio]{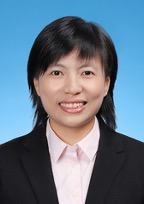
}}]{Xin Jin} (Senior Member, IEEE) received the M.S.
degree in communication and information system
and the Ph.D. degree in information and communication
engineering from the Huazhong University of
Science and Technology, Wuhan, China, in 2002 and
2005, respectively. From 2006 to 2008, she was a
Post-Doctoral Fellow with The Chinese University
of Hong Kong, Hong Kong. From 2008 to 2012,
she was a Visiting Lecturer with Waseda University,
Fukuoka, Japan. Since March 2012, she has
been with Shenzhen International Graduate School,
Tsinghua University, Beijing, China, where she is currently a Professor.
She is also a Distinguished Professor of the Peng Cheng Scholar. She
has authored or co-authored more than 200 conference and journal papers.
Her research interests include computational imaging, and power-constrained video processing and compression. She is a member of SPIE and ACM. She was a recipient of the Gold Medal of International Exhibition of Inventions of Geneva in 2024 and 2022, the Second Prize of the National Science and Technology Progress Award in 2016, the First Prize of Guangdong Science and Technology Award in 2015, and the ISOCC Best Paper Award in 2010.

\end{IEEEbiography}
\vspace{-38pt}
\begin{IEEEbiography}
[{\includegraphics[width=1in,height=1.25in,clip,keepaspectratio]{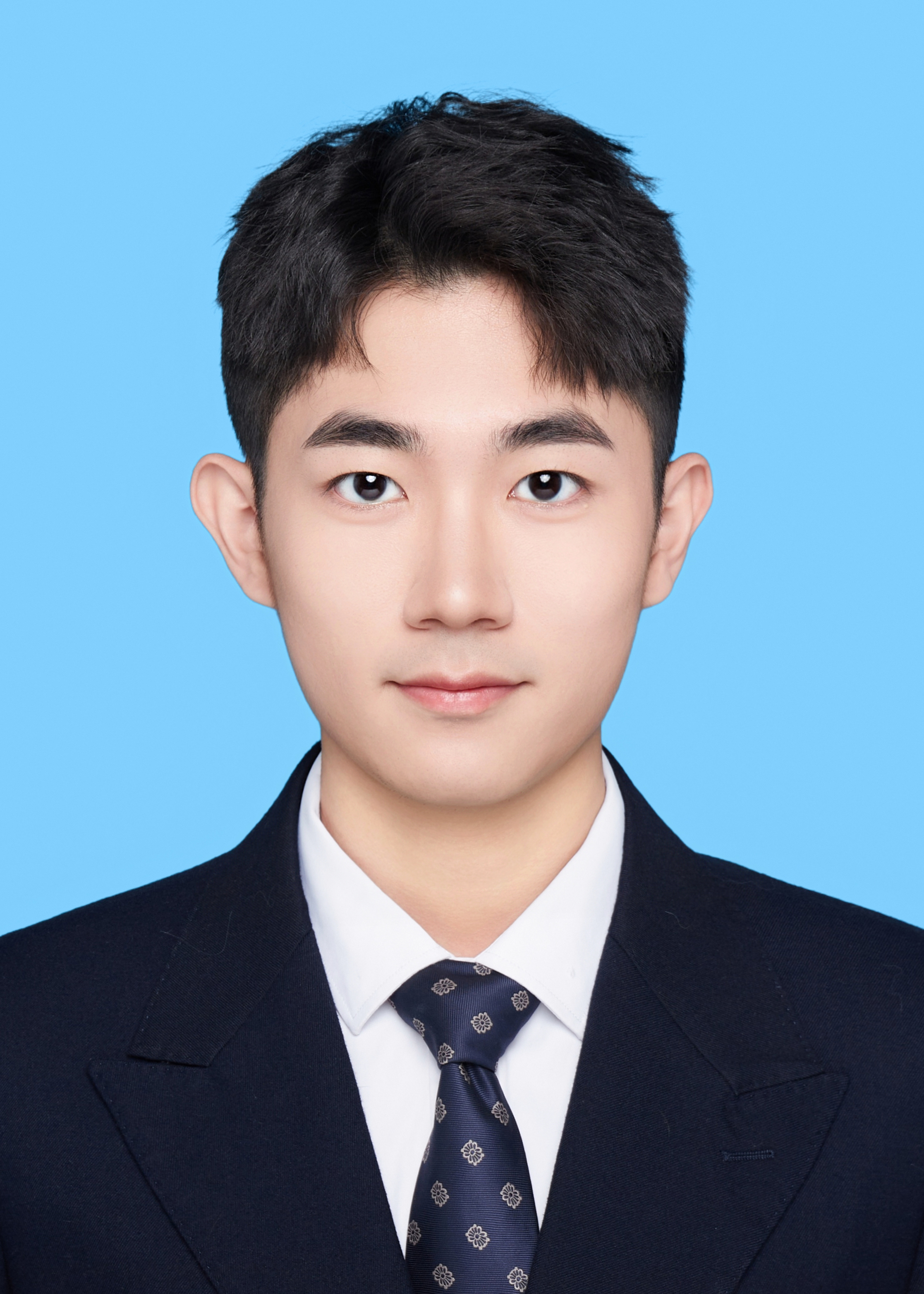}}]{Mingyu Liu} is currently pursuing the Ph.D. degree with Tsinghua Shenzhen International Graduate School, Tsinghua University. His research interests include the development of novel systems and algorithms for solving problems in multimodal perception, multimodal data fusion, and multimodal imaging.
\end{IEEEbiography}
\vspace{-38pt}
\begin{IEEEbiography}
[{\includegraphics[width=1in,height=1.25in,clip,keepaspectratio]{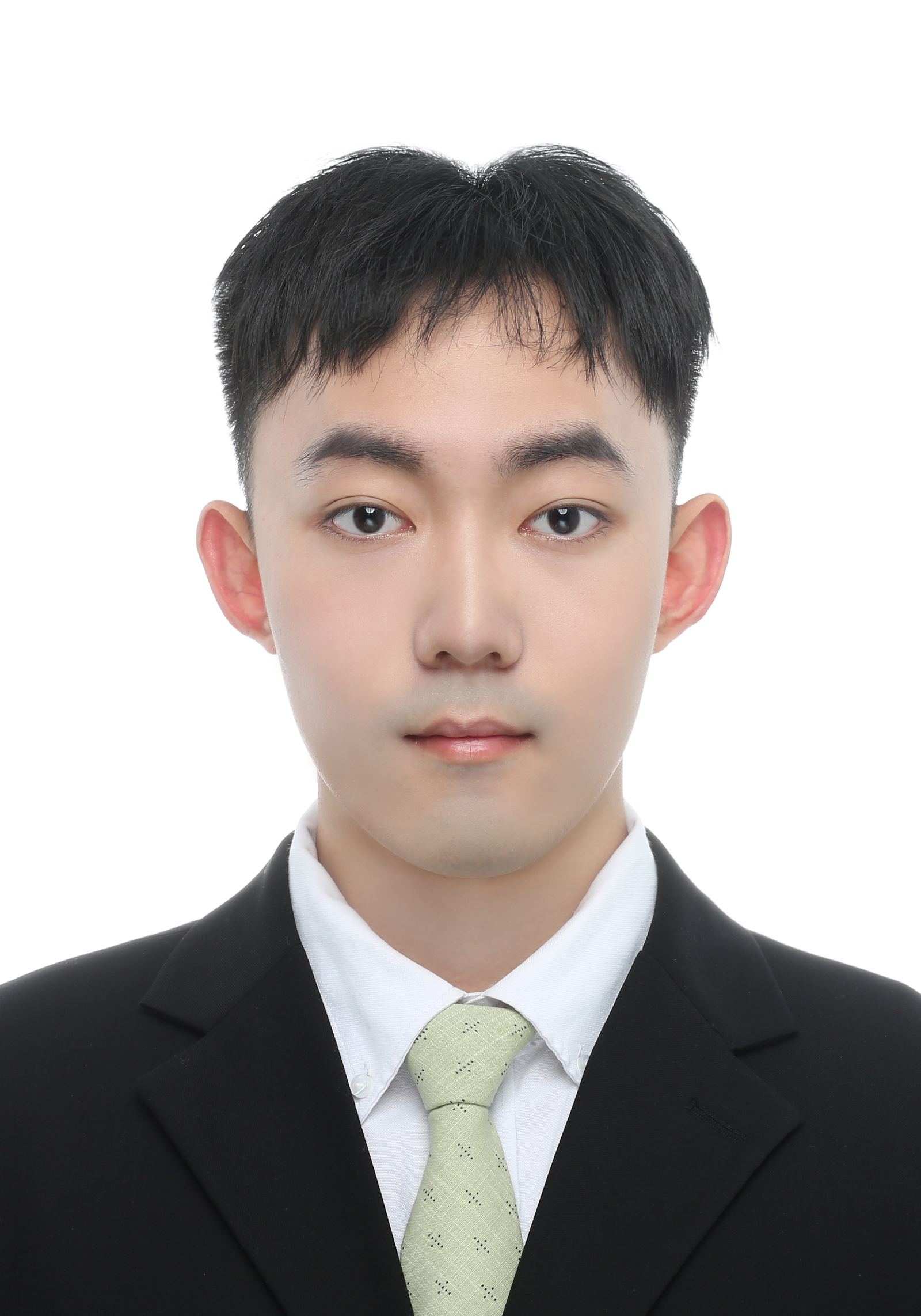}}]{Yihui Fan} is currently working toward the Ph.D. degree in the Control Science and Engineering with Shenzhen International Graduate School, Tsinghua University, China. His research interests include the development of new systems and algorithms for solving problems in light field sampling theory scattering imaging and light-field image stitching. He has published paper in IEEE TCSVT. He was the recipient of the Gold Medal of International Exhibition of Inventions of Geneva in 2024 and CITA Best Oral Presentation Award in 2023.
\end{IEEEbiography}

\end{document}